\pdfoutput=1

\documentclass[11pt]{article}

\usepackage{EMNLP2023}

\usepackage{times}
\usepackage{latexsym}
\usepackage{booktabs}
\usepackage[T1]{fontenc}

\usepackage[utf8]{inputenc}
\usepackage{graphicx}
\usepackage{enumitem}
\usepackage{multirow}
\usepackage{url}
\usepackage{tcolorbox}
\usepackage{arydshln}

\usepackage{microtype}

\usepackage{inconsolata}

\title{Chitranuvad: Adapting Multi-Lingual LLMs for Multimodal Translation}

\author{Shaharukh Khan, Ayush Tarun, Ali Faraz, Palash Kamble, Vivek Dahiya, \\  \textbf{Praveen Pokala *, Ashish Kulkarni *, Chandra Khatri *, } \\
\textbf{Abhinav Ravi * and Shubham Agarwal * }\\ \\
Krutrim AI, Bangalore, India\\
\texttt{* Senior Contributors}\\
\textsuperscript{Contact: \{shaharukh.khan, abhinav.ravi, shubham.agarwal1\}@olakrutrim.com} 
}

\begin{document}
\maketitle
\begin{abstract}

In this work, we provide the system description of our submission as part of the \textit{English-to-Lowres Multimodal Translation Task} at the Workshop on Asian Translation (WAT2024). We introduce Chitranuvad, a multimodal model that effectively integrates Multilingual LLM and a vision module for Multimodal Translation. Our method uses a ViT image encoder to extract visual representations as visual token embeddings which are projected to the LLM space by an adapter layer and generates translation in an autoregressive fashion. We participated in all the three tracks (Image Captioning, Text-only and Multimodal translation tasks) for Indic languages (ie. English translation to Hindi, Bengali and Malyalam) and achieved SOTA results for Hindi in all of them on the Challenge set while remaining competitive for the other languages in the shared task.



\end{abstract}

\section{Introduction}

\begin{figure*}
    \centering
\includegraphics[width=0.9\linewidth]{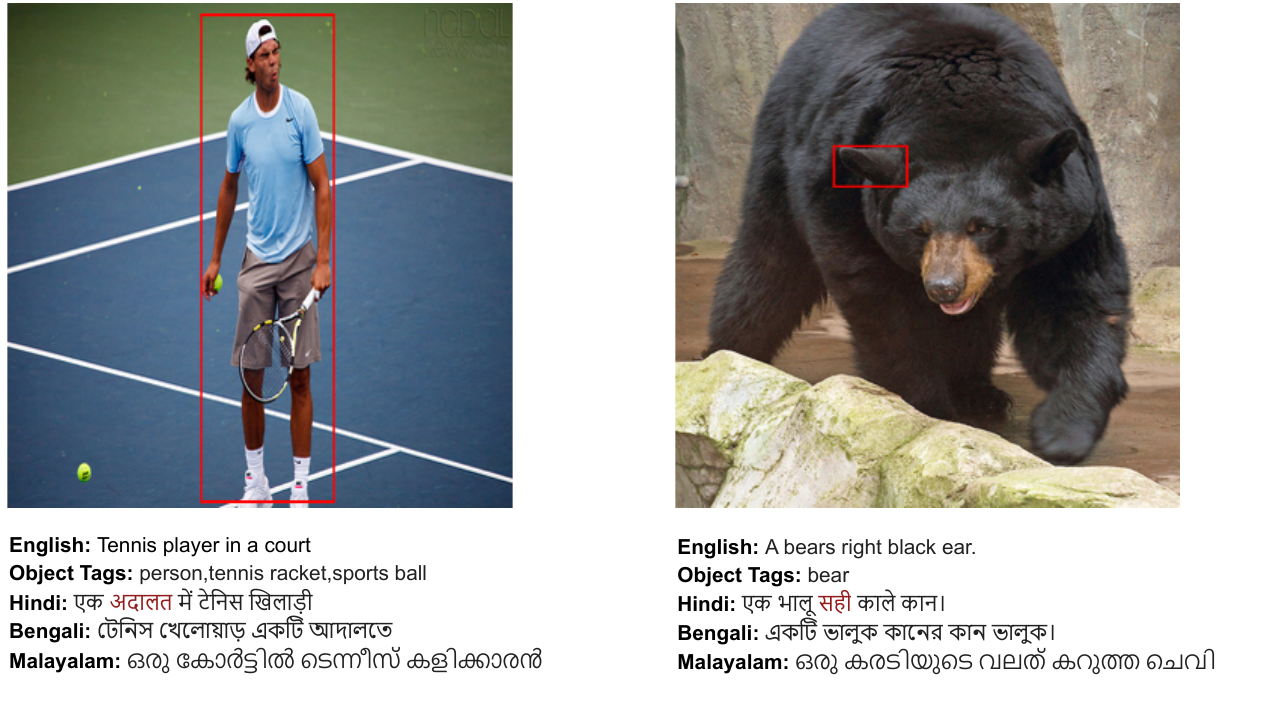}
    \caption{Multimodal Machine Translation task as part of English-to-lowres track where the source sentence is translated to multiple Indic languages (Hindi, Bengali, Malayalam) grounded in the image. Meaning of words like "court" and "right" in the translations can vary significantly depending on the visual context.}
    \label{fig:teaser}
\end{figure*}

Recently, there has been an increased interest in Multimodal Machine Translation (MMT) task~\cite{calixto-liu-2017-incorporating, delbrouck-dupont-2017-empirical, elliott-kadar-2017-imagination,yao2020multimodal} which involves translation between language pairs, incorporating other modalities (like images) as an auxiliary information. The visual cues act as `symbol grounding'~\cite{fodor1975language,harnad1990symbol,harnad2003minds,harnad2005cognize}, helping to resolve ambiguities in language \citep{rainie2012photos,hu2014we,specia2016shared,van2019task,caglayan2020simultaneous} by learning to connect language and perception~\cite{mooney2008learning,bisk2020experience}. For example, in order to correctly translate the word \textit{court} in Figure \ref{fig:teaser}, the model has to infer from the image that the statement is about tennis court and not the court as government institution.



Prior works mostly focused on translation from English to European languages ~\cite{elliott2016multi30k,specia2016shared} while the Indic languages remain largely unexplored, with an exception of the MMT shared task at the Workshop on Asian Translation (WAT) ~\cite{nakazawa2019proceedings,wat-2020-asian,wat-2021-asian,wat-2022-asian,wat-2023-asian}. 

The \textit{English-to-Lowres Multimodal Translation Task} at WAT-2024 targets the MMT task for three Indic medium-to-low-resource languages (Hindi, Bengali, Malayalam) and a low-resource African language Hausa. To assess the importance of the image modality, the task has been decoupled into three tracks: \textbf{1).} \textit{Text-only translation} where the source image is not used,  \textbf{2).} \textit{Image Captioning} where English source text is not used and \textbf{3).} \textit{Multimodal translation} which uses both the image and the text. We participated in all the three tracks for Indic languages only (Hindi, Bengali, Malayalam) under a non-constrained and proprietary multi-lingual and multimodal Large Language Model (LLM): \textit{Chitranuvad}\footnote{Chitranuvad literally means Image Translate in Hindi}.


In this paper, we provide a description of our multimodal LLM where we leverage a multi-lingual LLM backbone Krutrim~\cite{team2024krutrim}, coupled with a visual image encoder. Our contributions could thus be summarized as follows:

\begin{itemize}[noitemsep]
    \item We introduce Chitranuvad, a Large Multimodal model, adapted for multi-lingual translation, which leverages images and language modalities to provide an image grounded translation of the English sentence in the target Indic languages. 
    \item We showcase the effectiveness of task specific finetuning on the Visual Genome translation datasets and achieve SOTA performance.   
    \item We evaluate Chitranuvad and prior baselines on the English-to-Lowres Multimodal Translation Task and demonstrate the ability of our model to perform grounded translation, using different training strategies and ablations.   
    
\end{itemize}

The rest of the paper is organized as: Section \ref{sec:related-work} presents related research on multimodal machine translation while Section \ref{sec:model} explains our Chitranuvad model recipe in detail. We present the datasets used in Section \ref{sec:dataset}, followed by experimental findings in Section \ref{sec:experiments} and conclusion in Section \ref{sec:conclusion}.

\begin{figure*}
    \centering
\includegraphics[width=0.82\linewidth]{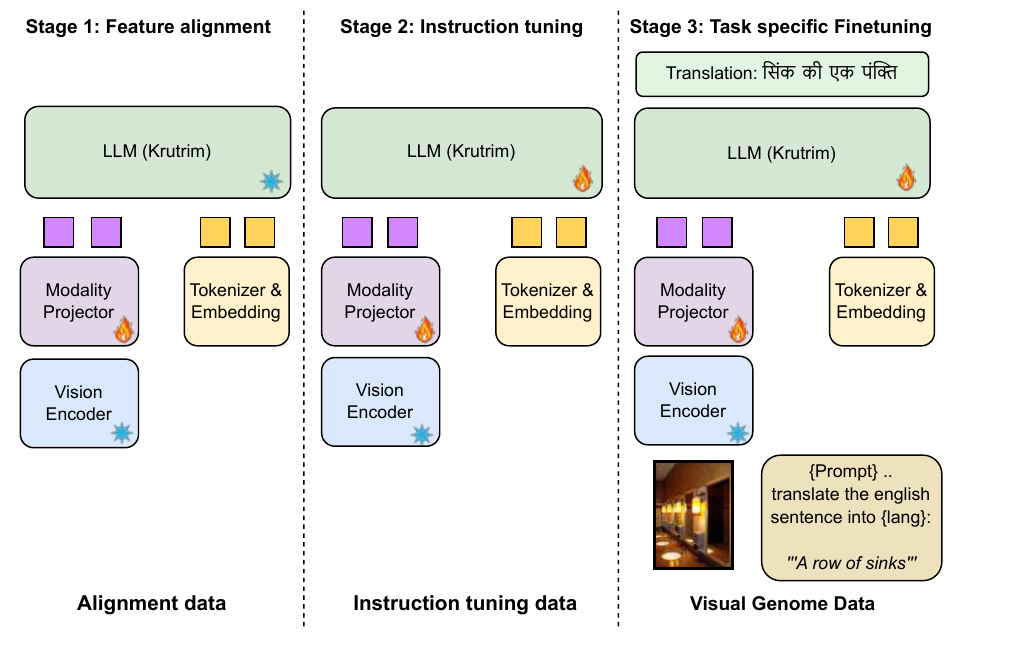} \caption{Chitranuvad model architecture with the three stage training pipeline described in Section \ref{sec:model}.}
    \label{fig:model}
\end{figure*}
\section{Related Work}
\label{sec:related-work}

Early Neural Machine Translation (NMT) and Image captioning systems~\cite{show2015tell,gao2018image} were based on Recurrent Neural Networks (RNNs) and their variants ~\cite{cho-etal-2014-learning, sutskever2014sequence,cho2014learning,hochreiter1997long}, often incorporating attention mechanisms ~\cite{bahdanau2014neural}. The seminal work of transformers~\cite{vaswani2017attention} paved the way for the development of high-quality image captioning ~\cite{chen2021captioning} as well as translation systems ~\cite{lewis2019bart}, even for low-resource languages~\cite{dabre2021indicbart,gala2023indictrans2}. Multimodal Machine Translation (MMT) systems witnessed a similar shift in their approrach ~\cite{caglayan2016multimodal,yao2020multimodal,guo2023layer}. 
Prior submissions to the MMT task at Workshop on Asian Translation~\cite{gain2021iitp, gupta2021vita, parida2022silo, dash2023bits, shahid2023odiagenai} also fall in this category. 

The next generation of Multimodal LLMs~\cite{lu2024deepseek,laurenccon2024matters,tong2024cambrian,xue2024xgen} can handle a variety of complex tasks, including machine translation and captioning, by utilizing cutting-edge architectures as an unified general purpose agent. These models often rely on pre-trained LLMs, with an exception of few, which train the models from scratch ~\cite{team2024chameleon,lu2024unified}. Most of these Vision Language Models (VLMs) follow the architecture of \cite{liu2024visual} where a CLIP \cite{radford2021learning} or a similar encoder is used to encode the image and projected into LLM's representation space using an adapter layer. Notably, \citet{wang2023cogvlm} offers a departure from conventional architectures by using distinct matrices and Feed Forward Networks for image modalities. Recent developments replace the image encoder with SigLIP \cite{zhai2023sigmoid} and the single-layer MLP projector with attention-based pooling \cite{laurenccon2024matters}. 

Advanced backbone LLMs ~\cite{brown2020language,touvron2023llama,achiam2023gpt,team2023gemini,jiang2024mixtral, team2024gemma} however have a primary focus for English and European languages. There have been relatively few LLMs for Indic languages, such as Airavata \cite{gala2024airavata},  Navarsa \cite{NavarasaTeluguLLMLabs}, Kannada LLaMA, Tamil LLaMA \cite{balachandran2023tamilllama}, Odia LLaMA \cite{kohli2023building}, to name a few. However, most of these LLMs are an extension and finetuned version of LLaMA/Gemma for Indic languages, which don't fully capture the nuances of the language. This could be attributed to the fact that Indic languages are under-represented in Common Crawl (which majorly forms the training corpus of LLMs), despite India constituting 18\% of the global population. Hindi, for example, does not show-up in the top 20 languages despite being the 3rd most spoken ~\cite{buck2014n,penedo2023refinedweb}. Closed-source models such as Krutrim~\cite{team2024krutrim} and Sutra~\cite{bendale2024sutra} represent exceptions, as they are trained from scratch. Currently, PALO~\cite{maaz2024palo} is a multimodal LLM that supports only Hindi and Bengali. However, to the best of our knowledge, there are no other open-source multimodal LLMs trained specifically for low-resource Indic languages. In contrast, we developed a multilingual multimodal translation system that supports 10 Indic languages based on our general purpose multimodal model Chitrarth \cite{khan2025chitrarth}.
%

\section{Model and Training Recipes}
\label{sec:model}

Figure \ref{fig:model} provides an overview of our architecture and the multi-stage training pipeline. Our \textit{Chitranuvad} model architecture borrows heavily from LLaVA-like models~\cite{liu2024visual,liu2024improved}, where we use pre-trained Krutrim LLM~\cite{team2024krutrim} instead, as the autoregressive multi-lingual LLM backbone. Our Krutrim LLM is trained across 10 languages and natively supports all the 3 Indic languages (Hindi, Bengali, Malayalam) used as part of the shared task.

For the multimodal training, we first encode images through a vision encoder. Next, the modality projection (adapter/connector) layer projects the vision embeddings into the LLM embedding space, creating a sequence of visual tokens. The multi-lingual LLM then generates the response conditioned on these visual embedding tokens. The Krutrim LLM model supports a context length of 4096 tokens, out of which 576 tokens are used for the image representation, obtained after the modality projector layer. For the projection layer, we experiment with both single layer projection~\cite{liu2023visualinstructiontuning} as well as a two-layer MLP vision-language connector with non-linearity~\cite{liu2024visual}. 
We also experiment with pre-trained CLIP ViT-L/14@336px~\cite{radford2021learning} as well as SigLIP-SO400M \cite{zhai2023sigmoidlosslanguageimage} for the vision encoder. Similar to the LLaVA model, we generate multi-turn conversational data for instruction tuning our model, which we expand upon in Section \ref{sec:dataset}. We train our model in multiple stages:

\textbf{Stage 1: Pre-Training (PT) for Feature Alignment.} In this stage, we do the pre-training with image-text pairs, where the projector layer is trained while the vision encoder and LLM is kept frozen. Here, each sample is treated as a single-turn conversational instruction tuning data. 

\textbf{Stage 2: Instruction Tuning.} Similar to LLaVA models ~\cite{liu2023visualinstructiontuning,liu2024visual}, we also keep the vision encoder frozen during the second stage of training. However, here we also update the LLM weights apart from tuning the modality projection layer. This stage aims to build a general purpose Multimodal agent (chatbot) which can follow complex instructions across multiple-turns of the conversation. We focus on developing a specialized multimodal translation system in the Stage 3.

\textbf{Stage 3: Task-specific Fine-Tuning.} We follow a similar recipe to that of Stage 2 for the (Machine Translation) task-specific fine-tuning and update weights for the projection layer and the LLM while keeping the vision encoder frozen. Here, we experiment with both LoRA style training ~\cite{hu2021lora,houlsby2019parameter} as well as full parameter fine-tuning on the shared task translation data. 



\begin{table}
\centering
\resizebox{0.99\linewidth}{!}{
\begin{tabular}{lccccc}
\toprule
\textbf{Split} & \textbf{\#Instances} &
\textbf{English} & \textbf{Hindi} & \textbf{Bengali} & \textbf{Malayalam} \\
\midrule
Train & 28930 & 5.09 & 5.13 & 4.07 & 3.86 \\
Valid & 998 & 5.08 & 5.04 & 4.06 & 3.75 \\
Test & 1595 & 5.07 & 4.95 & 4.14 & 3.76 \\
Challenge & 1400 & 6.04 & 6.35 & 4.92 & 4.48 \\ 
\bottomrule
\end{tabular}}
\caption{Total number of instances and average number of tokens for the text in English and splits of different Visual Genome datasets in other languages.}
\label{tab:visgenome-stats}


\end{table}

\begin{table}
    \centering
    \begin{tcolorbox}
\footnotesize
\textbf{Multimodal Translation}:

\textit{Human:} You are given an image  and coordinates of a bounding box as: x1=\{x1\}, y1=\{y1\}, x2=\{x1+x2\}, y2=\{y1+y2\}. Using the context of the objects or items available in the bounding box translate the following sentence from English into \{lang\} language. You are also provided labels of the objects in the image as: \{labels\}. English sentence is: \{sentence\}. 

\textit{System:} \{translation\}.

\textbf{Text only translation}:

\textit{Human:} Translate the following sentence from English into \{lang\} language. English sentence is: \{sentence\}. 

\textit{System:} \{translation\}.

\textbf{Image captioning}:
\textit{Human:} You are given an image  and coordinates of a bounding box as: x1=\{x1\}, y1=\{y1\}, x2=\{x1+x2\}, y2=\{y1+y2\}. You are also provided labels of the objects in the image as: \{labels\}. Provide a short caption of the object in \{lang\} language.

\textit{System:} \{caption\}.

    \end{tcolorbox}
    \caption{Different prompt templates  for creating task specific fine-tuning data, used in Stage 3 training.}
    \label{tab:data-prompts}
\end{table}

\section{Dataset}
\label{sec:dataset}
In this section, we describe the data resources utilized throughout our experiments. 

\textbf{Stage 1:} 
In our initial experiments, we use the LLaVA-Pretrain-LCS-558K data for pre-training our model in Stage 1. However, recent works~\cite{tong2024cambrian} showed that more adapter data is beneficial for the model, such as the 1.2M ShareGPT4V-PT \cite{chen2023sharegpt4v} image-captioning dataset, which we use in Stage 1 training. 
We also translated this data in the 10 Indic languages that our LLM natively supports, using an in-house text Machine Translation system. 
We sample translations across different languages (including English) in an equal ratio and ensure that PT data limits to 1.2M data points in our final data mix.




\textbf{Stage 2:} For the second stage instruction tuning, eliciting visual reasoning abilities, we experiment with both LLaVA-Instruct-150K~\cite{liu2023visualinstructiontuning} and LLaVA-1.5-665K~\cite{liu2024improved} where we find continued improvements with the 665K version. Similar to pre-training data, we also translated the LLaVA-1.5-665K into multiple languages. 
Recently released Cauldron dataset~\cite{laurenccon2024matters} is a collection of 50 academic Vision-language tasks. In our final submission, we also include the translated versions and the original English language based Cauldron apart from the proprietary multi-modal dataset in the training mix. It must be noted that the English only Visual Genome might be a part of this stage's training data through various academic datasets, though not for the translation task.  
\begin{table*}[htbp!]
    \centering
    \resizebox{0.99\linewidth}{!}{
    \setlength{\tabcolsep}{3pt} 
    \begin{tabular}{lcc|cc|cc|cc|cc|cc} 
        \toprule
        & \multicolumn{2}{c}{\textbf{Hi-Ch}} 
        & \multicolumn{2}{c}{\textbf{Hi-Test}} 
        & \multicolumn{2}{c}{\textbf{Bn-Ch}} 
        & \multicolumn{2}{c}{\textbf{Bn-Test}}
        & \multicolumn{2}{c}{\textbf{Ml-Ch}} 
        & \multicolumn{2}{c}{\textbf{Ml-Test}}        
        \\
         \cmidrule(lr){2-3} \cmidrule(lr){4-5}
        \cmidrule(lr){6-7} \cmidrule(lr){8-9}
        \cmidrule(lr){10-11} \cmidrule(lr){12-13}

        \textbf{Submission}  & \textbf{BLEU $\uparrow$} & \textbf{RIBES $\uparrow$}
         & \textbf{BLEU $\uparrow$} & \textbf{RIBES $\uparrow$} 
         & \textbf{BLEU $\uparrow$} & \textbf{RIBES $\uparrow$}
         & \textbf{BLEU $\uparrow$} & \textbf{RIBES $\uparrow$} 
         & \textbf{BLEU $\uparrow$} & \textbf{RIBES $\uparrow$}
         & \textbf{BLEU $\uparrow$} & \textbf{RIBES $\uparrow$}         \\
        \midrule
        SILO NLP	 & 29.6 & 0.73 &  36.2 & 0.79 & 22.6 & 0.61 & 41.0 & 0.77 & 14.6 & 0.39 & 30.8 & 0.60 \\
        NLP Voices	 & 41.8 & 0.81 &  43.1 & 0.82 & 32.9 & 0.71 & 39.8 & 0.75 & 19.6 & 0.54 & 30.6 & 0.64 \\
        Volta & 51.7 & 0.86 & 44.1 & 0.82 & - & - & - & - & - & - & - & - \\
        ODIAGEN & 53.6 & 0.86 &  44.6 & 0.83 & \textbf{47.8} & \textbf{0.82} & \textbf{49.2} & \textbf{0.8} & 39.7 & 0.75 & 46.6 & 0.75 \\
        \hdashline
        Ours (leaderboard) & \textbf{54.1} & \textbf{0.86} & 43.3 & 0.81 & 44.2 & 0.79 & 45.1 & 0.77 & 34.0 & 0.65 & 37.8 & 0.63 \\

        Ours\textdagger & \textbf{55.3} & \textbf{0.87} & \textbf{44.7} & \textbf{0.83} & \underline{46.7} & \underline{0.81} & \underline{48.1} & \underline{0.79} & \textbf{40.6} & \textbf{0.75} & \textbf{51.7} & \textbf{0.88} \\

        \bottomrule
    \end{tabular}
    }
    \caption{English-to-lowres leaderboard scores for Text-only task for Indic languages (Team 007). In the following tables, \textdagger denotes the results after submission deadline using the IndicTrans2 evaluation scripts, all the other results are reported using the shared task dashboard.}
    \label{tab:leaderboard-txt}
\end{table*}

\begin{table*}[htbp]
    \centering
    \resizebox{0.99\linewidth}{!}{
    \setlength{\tabcolsep}{3pt} 
    \begin{tabular}{lcc|cc|cc|cc|cc|cc} 
        \toprule
        & \multicolumn{2}{c}{\textbf{Hi-Ch}} 
        & \multicolumn{2}{c}{\textbf{Hi-Test}} 
        & \multicolumn{2}{c}{\textbf{Bn-Ch}} 
        & \multicolumn{2}{c}{\textbf{Bn-Test}}
        & \multicolumn{2}{c}{\textbf{Ml-Ch}} 
        & \multicolumn{2}{c}{\textbf{Ml-Test}}        
        \\
         \cmidrule(lr){2-3} \cmidrule(lr){4-5}
        \cmidrule(lr){6-7} \cmidrule(lr){8-9}
        \cmidrule(lr){10-11} \cmidrule(lr){12-13}

        \textbf{Method}  & \textbf{BLEU $\uparrow$} & \textbf{RIBES $\uparrow$}
         & \textbf{BLEU $\uparrow$} & \textbf{RIBES $\uparrow$} 
         & \textbf{BLEU $\uparrow$} & \textbf{RIBES $\uparrow$}
         & \textbf{BLEU $\uparrow$} & \textbf{RIBES $\uparrow$} 
         & \textbf{BLEU $\uparrow$} & \textbf{RIBES $\uparrow$}
         & \textbf{BLEU $\uparrow$} & \textbf{RIBES $\uparrow$}         \\
        \midrule
        ODIANLP & 0 & 0.04 & 0.8 & 0.06 & - & - & - & - & - & - & - & - \\

        NLPHUT & - & - & - & - & - & - & - & - & 0.9 & 0.02 & 0.9 & 0.05 \\
        \hdashline
        Ours (leaderboard) & \textbf{1.3} & \textbf{0.13} & \textbf{2.8} & \textbf{0.18} & \textbf{0.4} & \textbf{0.04} & \textbf{1.8} & \textbf{0.11} & 0.3 & 0.04 & 0.9 & \textbf{0.06} \\
        
        \bottomrule
    \end{tabular}
    }
    \caption{English-to-lowres leaderboard scores for Image captioning track. Ours is the only multi-lingual model which can handle all the 3 Indic languages for image captioning.}
    \label{tab:leaderboard-cap}
\end{table*}

\textbf{Stage 3:} 
For the Stage 3, we work with the aligned multi-lingual Visual Genome~\cite{krishna2017visual} datasets, i.e.\ Hindi \cite{hindi-visual-genome:2019}, Bengali \cite{sen2022bengali-visgen} and   Malayalam \cite{11234/1-3533-malayalam-visgen}, bundled as part of the shared task. Each row in the dataset consists of the following fields: \textit{i).} MS COCO ~\cite{lin2014microsoft} image id \textit{ii).} English utterance 
\textit{iii).} Translated utterance in Hindi/ Bengali/ Malayalam \textit{iv).} Bounding box of the area in the image that the utterance is based on. While there is also a track for Hausa language~\cite{abdulmumin2022hausa}, we don't include this in our training data. Table~\ref{tab:visgenome-stats} provides the statistics of the different versions of the dataset, which we transform into instruct tuning format, similar to Stage 1 and 2 data (see Table \ref{tab:data-prompts}). To increase the efficacy of our model, we enrich the dataset with the labels of different objects in the image (object tags), similar to ~\cite{gupta2021vita}. We use SOTA (state-of-the-art) YOLOv8~\cite{varghese2024yolov8} for object detection compared to the prior works, which relied on Faster R-CNN models~\cite{wu2019detectron2,girshick2015fast}. We also calculate the Intersection-over-union (IoU) for the detected and the dataset provided bounding boxes to get the most relevant object tag. However, we found a decreased performance against including the labels of all the detected objects.

\section{Experimental Results and Discussion}
\label{sec:experiments}

This section details our experimental setup and presents the results of our comparative studies. 


\subsection{Implementation}

We use HuggingFace Transformers~\cite{wolf2019huggingface} based on PyTorch~\cite{paszke2019pytorch} for our experiments. We consider PALO~\cite{maaz2024palo} as a multi-lingual multi-modal baseline and use the code provided with the repository\footnote{\url{https://github.com/mbzuai-oryx/PALO}}. The shared task provides a leaderboard based on the automatics metrics of BLEU~\cite{papineni2002bleu} and RIBES~\cite{isozaki2010automatic}.
For reporting BLEU, we used the evaluation scripts\footnote{\url{https://github.com/AI4Bharat/IndicTrans2}} provided with ~\cite{gala2023indictrans} and the official repository for RIBES\footnote{\url{https://github.com/nttcslab-nlp/RIBES}}. Similar to previous works~\cite{gupta2021vita}, we also report the results after tokenizing the outputs using indic-tokenizer\footnote{\url{https://github.com/ltrc/indic-tokenizer}}. Our Stage 1 and Stage 2 tuning follow similar hyperparameters as the LLaVA model~\cite{liu2023visualinstructiontuning} unless specified otherwise. 
For Stage 3 fine-tuning, we conducted multiple experiments for hyperparameter search of learning rate (1e-3, 1e-4, and 1e-5); as well as multiple epochs (1, 2, 3, and 5). We observed rapid over fitting after only one epoch while a learning rate of 1e-4 yielded the highest overall performance. All our further experiments are reported based on this configuration.  

\begin{table*}[htbp!]
    \centering
    \resizebox{0.99\linewidth}{!}{
    \setlength{\tabcolsep}{3pt} 
    \begin{tabular}{lcc|cc|cc|cc|cc|cc} 
        \toprule
        & \multicolumn{2}{c}{\textbf{Hi-Ch}} 
        & \multicolumn{2}{c}{\textbf{Hi-Test}} 
        & \multicolumn{2}{c}{\textbf{Bn-Ch}} 
        & \multicolumn{2}{c}{\textbf{Bn-Test}}
        & \multicolumn{2}{c}{\textbf{Ml-Ch}} 
        & \multicolumn{2}{c}{\textbf{Ml-Test}}        
        \\
         \cmidrule(lr){2-3} \cmidrule(lr){4-5}
        \cmidrule(lr){6-7} \cmidrule(lr){8-9}
        \cmidrule(lr){10-11} \cmidrule(lr){12-13}

        \textbf{Submission}  & \textbf{BLEU $\uparrow$} & \textbf{RIBES $\uparrow$}
         & \textbf{BLEU $\uparrow$} & \textbf{RIBES $\uparrow$} 
         & \textbf{BLEU $\uparrow$} & \textbf{RIBES $\uparrow$}
         & \textbf{BLEU $\uparrow$} & \textbf{RIBES $\uparrow$} 
         & \textbf{BLEU $\uparrow$} & \textbf{RIBES $\uparrow$}
         & \textbf{BLEU $\uparrow$} & \textbf{RIBES $\uparrow$}         \\
        \midrule
        IIT-P	 & 37.5 & 0.79 &  42.5 & 0.81 & - & - & - & - & - & - & - & - \\
        ODIAGEN	 & 42.8 & 0.82 &  41.6 & 0.81 & 30.5 & 0.69 & 42.4 & 0.76 & - & - & - & - \\
        Volta & 51.6 & 0.86 & 44.6 & 0.82 & - & - & - & - & - & - & - & - \\
        BITS-P & 52.1 & 0.85 & 45.0 & 0.83 & 48.7 & 0.83 & 50.6 & 0.81 & 42.2 & 0.76 & 51.9 & 0.80 \\
        \hdashline
        Ours (leaderboard) & \textbf{53.4} & \textbf{0.842} &  43.7 & 0.81 & 44.8 & 0.78 & 44.5 & 0.76 & 39.8 & 0.74 & \textbf{51.9} & 0.78 \\
        Ours\textdagger & \textbf{54.7} & \textbf{0.86} &  \underline{43.9} & \textbf{0.83} & \underline{46.9} & \underline{0.81} & \underline{47.7} & \underline{0.79} & \underline{40.3} & \underline{0.74} & \textbf{51.9} & \textbf{0.93} \\
        \bottomrule
    \end{tabular}
    }
    \caption{English-to-lowres leaderboard Scores for Multimodal translation track across multiple languages (Team 007). \textdagger denotes the results after submission deadline using the IndicTrans2 evaluation scripts}
    \label{tab:leaderboard-mmt}
\end{table*}

\subsection{Results for different tracks}

\begin{figure*}
    \centering
\includegraphics[width=\linewidth]{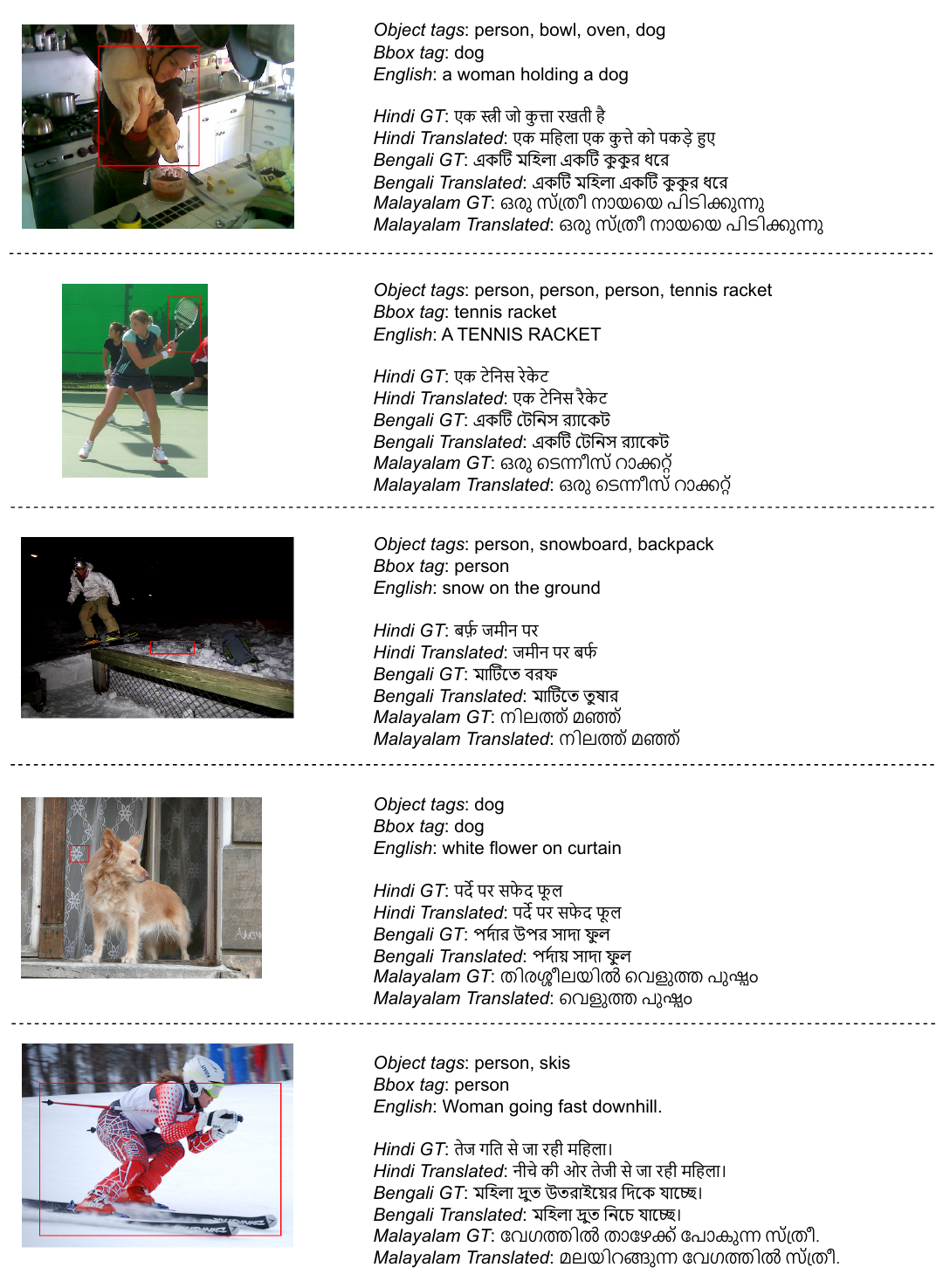}
    \caption{English-to-lowres Multimodal Machine Translation track supports translation of source sentence into multiple Indic languages (Hindi, Bengali, Malayalam). We enrich the dataset to include labels of all the identified objects. We show the outputs of our best model which is trained with a mix of multi-lingual data in all the 3 stages.}
    \label{fig:results}
\end{figure*}

Table \ref{tab:leaderboard-txt}, \ref{tab:leaderboard-cap} and \ref{tab:leaderboard-mmt} present the results for text-only, image captioning and the Multimodal translation task respectively. For the text-only task, our Chitranuvad model was trained with image data till Stage 2. In Stage 3, we only finetune with text only translations. During inference, we prompt the model with text only translations and dont provide images. Our model achieves SOTA on Hindi and Malayalam Challenge and Test sets while being competitive on the Bengali dataset (see Table \ref{tab:leaderboard-txt}). We were the only submission which could do image captioning in all the 3 languages (see Table \ref{tab:leaderboard-cap}). For the MMT task, we achieved SOTA on Hindi Challenge and Malayalam test set while being competitive on the other languages. We also provide cherry-picked system outputs of our best Multimodal LLM in Table ~\ref{fig:results}. From our manual inspection, we saw that our generated translations are better than the ground truth. For example, in the last snippet, our model correctly translates the word `downhill', which the gold translation fails to capture. 

\begin{table*}[htbp!]
    \centering
    \resizebox{0.99\linewidth}{!}{
    \setlength{\tabcolsep}{3pt} 
    \begin{tabular}{lcc|cc|cc|cc|cc|cc} 
        \toprule
        & \multicolumn{2}{c}{\textbf{Hi-Ch}} 
        & \multicolumn{2}{c}{\textbf{Hi-Test}} 
        & \multicolumn{2}{c}{\textbf{Bn-Ch}} 
        & \multicolumn{2}{c}{\textbf{Bn-Test}}
        & \multicolumn{2}{c}{\textbf{Ml-Ch}} 
        & \multicolumn{2}{c}{\textbf{Ml-Test}}        
        \\
         \cmidrule(lr){2-3} \cmidrule(lr){4-5}
        \cmidrule(lr){6-7} \cmidrule(lr){8-9}
        \cmidrule(lr){10-11} \cmidrule(lr){12-13}

        \textbf{Method}  & \textbf{BLEU $\uparrow$} & \textbf{RIBES $\uparrow$}
         & \textbf{BLEU $\uparrow$} & \textbf{RIBES $\uparrow$} 
         & \textbf{BLEU $\uparrow$} & \textbf{RIBES $\uparrow$}
         & \textbf{BLEU $\uparrow$} & \textbf{RIBES $\uparrow$} 
         & \textbf{BLEU $\uparrow$} & \textbf{RIBES $\uparrow$}
         & \textbf{BLEU $\uparrow$} & \textbf{RIBES $\uparrow$}         \\
        \midrule
        PALO-7B  & 14.8 & 0.585 & 13.3 & 0.567 & 7.9 & 0.469 & 9.6 & 0.464   & 0.1 & 0.001 & 0 & 0  \\
        PALO-13B  & 15.8 & 0.605 & 14.9 & 0.605 & 6.7 & 0.44 & 7.0 & 0.45 & 0.1 & 0.004 & 0 & 0 \\
        \hdashline
        Chitranuvad (Eng) & 18.3 & 0.629 & 12.9 & 0.585 & 8.7 & 0.512 & 8.3 & 0.477 & 8.7 & 0.487 & 7.3 & 0.426 \\
        Chitranuvad (Eng+Hindi) & 20.0 & \textbf{0.698} & 14.8 & 0.653 & 9.4 & 0.537 & 8.9 & 0.494 & 9.2 & 0.511 & 8.6 & 0.466 \\
        Chitranuvad (Multilingual) & \textbf{25.0} & 0.694 & \textbf{19.0} & \textbf{0.66} & \textbf{11.4} & \textbf{0.569} & \textbf{9.7} & \textbf{0.515} & \textbf{12.2} & \textbf{0.54} & \textbf{10.3} & \textbf{0.486} \\         
        \bottomrule
    \end{tabular}
    }
    \caption{0-shot results for Multimodal Machine Translation track as discussed in Section \ref{subsec:0-shot}. \textit{Eng} denotes only English data is used in Stage 1 and 2. \textit{Eng+Hindi} denotes English and Hindi data in Stage 2. As expected, we find the best scores when the training data mix consists of data from the 10 Indic languages.}
    \label{tab:0-shot}
\end{table*}

\subsection{0-shot on the Shared task data}
\label{subsec:0-shot}

We evaluate the efficacy of our model after Stage 2 as the 0-shot setting, where we don't fine-tune specifically for the shared task translation data. In our preliminary experiments, we only use the English versions of the datasets mentioned in Section \ref{sec:dataset} for both Adapter tuning (Stage 1) and Instruction tuning (Stage 2). Exceptionally, our Krutrim LLM still retained multi-lingual capabilities, evident from the scores in Table \ref{tab:0-shot}. When we also include Hindi data in the training mix, we find an uplift on the Hindi translation task. Including multi-lingual data in both the stages further showed improvement on all three language translation tasks in the 0-shot setting. We thus use this as the base model in the Stage 3 training. We also evaluate against the open-source baseline of PALO-7/13B ~\cite{maaz2024palo} in the 0-shot setting. To our surprise, our Chitranuvad model consistently outperforms the 0-shot PALO baseline, even when our model is fine-tuned with English only data in both the stages. We hypothesis that this is because the base LLM Vicuna~\cite{zheng2024judging} used in PALO is not inherently multi-lingual in nature. 



\begin{table*}[htbp!]
    \centering
    \resizebox{0.99\linewidth}{!}{
    \setlength{\tabcolsep}{3pt} 
    \begin{tabular}{lcc|cc|cc|cc|cc|cc} 
        \toprule
        & \multicolumn{2}{c}{\textbf{Hi-Ch}} 
        & \multicolumn{2}{c}{\textbf{Hi-Test}} 
        & \multicolumn{2}{c}{\textbf{Bn-Ch}} 
        & \multicolumn{2}{c}{\textbf{Bn-Test}}
        & \multicolumn{2}{c}{\textbf{Ml-Ch}} 
        & \multicolumn{2}{c}{\textbf{Ml-Test}}        
        \\
         \cmidrule(lr){2-3} \cmidrule(lr){4-5}
        \cmidrule(lr){6-7} \cmidrule(lr){8-9}
        \cmidrule(lr){10-11} \cmidrule(lr){12-13}

        \textbf{Method}  & \textbf{BLEU $\uparrow$} & \textbf{RIBES $\uparrow$}
         & \textbf{BLEU $\uparrow$} & \textbf{RIBES $\uparrow$} 
         & \textbf{BLEU $\uparrow$} & \textbf{RIBES $\uparrow$}
         & \textbf{BLEU $\uparrow$} & \textbf{RIBES $\uparrow$} 
         & \textbf{BLEU $\uparrow$} & \textbf{RIBES $\uparrow$}
         & \textbf{BLEU $\uparrow$} & \textbf{RIBES $\uparrow$}         \\
        \midrule
        LoRA & 42.1 & 0.721 & 34.5 & 0.770 & 28.3 & 0.69 & 30.4 & 0.669 & 23.2 & 0.61 & 27.0 & 0.601 \\
        \hdashline
        Bi-lingual (Hi Stage 3) & 53.0 & 0.848 & 43.3 & 0.833 & 0.3 & 0.006 & 0.1 & 0.001 & 0.2 & 0.003 & 0 & 0 \\
        Bi-lingual (Bn Stage 3) & 0.3 & 0.005 & 0.1 & 0.001 & 45.4 & 0.797 & 46.6 & 0.781 & 0.3 & 0.003 & 0.1 & 0.001 \\
        \hdashline
        Only Stage 1, 3 & 53.6 & 0.853 & 43.4 & 0.826 & 45.2 & 0.801 & 46.4 & 0.788 & 38.2 & 0.735 & 50.3 & 0.781 \\
        \hdashline
        Back Translation mix & 53.8 & 0.856 & 43.6 & 0.828 &  46.0 & 0.806 &  46.8 & 0.792 & 37.5 & 0.729 &  46.3 & 0.738 \\
        
        \bottomrule
    \end{tabular}
    }
    \caption{Different finetuning strategies for Multimodal Machine Translation track as described in Section \ref{subsec:finetuning} in the order of discussion. Our Stage 3 full finetuning training performs the best compared to the other training recipes.}
    \label{tab:bad-results}
\end{table*}

\subsection{Other fine-tuning approaches}
\label{subsec:finetuning}

In this section, we elaborate on key findings with different fine-tuning approaches, with all the results reported in Table \ref{tab:bad-results}. 

\textbf{LoRA vs Full finetuning.} We investigated the efficacy of full fine-tuning versus Low-Rank Adaptation (LoRA) using Visual Genome data. Our experiments (see Table \ref{tab:bad-results}) reveal that full fine-tuning consistently outperforms LoRA, i.e.\ LoRA learns less~\cite{biderman2024lora}. 

\textbf{Bi-lingual vs Multi-lingual}
For Stage 3 training, we experiment with training specialized models for each language (Hindi and Bengali) compared to multi-lingual setting with a mix of data from all the three Indic languages. We didn't find any improvement over multi-lingual model but instead observe catastrophic forgetting~\cite{zhai2023investigating, tong2024cambrian}, where the translation abilities of the model in the other languages deteriorate completely. We hypothesize that a mix of multiple languages probably act as regularizaton and enhance the general translation capabilities. 

\textbf{Do we need second stage training?} Similar to ~\cite{tong2024cambrian}, we investigate if we even need Stage 2 instruction tuning. We find that our model, if finetuned directly on Visual Genome translation data (i.e.\ Stage 1 and 3 training only) performed comparable to the previous baselines. Including Stage 2 training provided an uplift in the scores with an added advantage of building a general purpose Multimodal agent.

\textbf{Back translation}
Back translations, i.e.\ using the reverse translations have been a popular technique both for data augmentation as well as post-processing or re-ranking techniques in traditional Machine Translation and Natural Language Generation systems~\cite{sennrich2015improving, li2015diversity, agarwal2018char2char, edunov2018understanding, graca-etal-2019-generalizing}. This involves re-translating content from the target language to its source language. Thus, apart from the original task of En -> Hi/Bn/Ml, we also included in our training corpus, the task of reverse translation from Hi/Bn/Ml -> En in the Stage 3 training mix. However, in our experiments, we found that this strategy showed decreased performance in terms of automatic metrics.

\section{Conclusion}
\label{sec:conclusion}
We present Chitranuvad, a multimodal LLM that is adapted for image grounded Machine translation. Our model encodes images using a pre-trained image encoder~\cite{alexey2020image} and translates the English sentences autoregressively into different Indic languages (Hindi, Bengali, Malayalam) using a pre-trained LLM.
Empirically, our model outperforms previous baselines for different tasks. 
However, we also observed that vision modality had little impact on the translation, echoing the observations from ~\cite{gronroos2018memad, lala2018sheffield, wu-etal-2021-good, li2022vision}.

\textbf{Broader Impact:} 
We believe our work paves way for building next generation assistants which can do multimodal machine translation. We believe these systems can empower different sectors like education, healthcare, banking and financial services, etc. to name a few. 

\textbf{Limitations and Future Work:} 
While this work is focused to three Indic languages (Hindi, Bengali, Malayalam), we consider our approach as a first step towards building general purpose multi-lingual system which can handle various Indic languages. While in our current setup, we freeze the vision encoder during training, recent works have shown that unfreezing the vision encoder with Perceiver Resampler~\cite{jaegle2021perceiver}, helps learn better representations~\cite{laurenccon2024matters,tong2024cambrian}, which we plan to explore in the future. 

\section*{Acknowledgements}
We thank Bhavish Aggarwal and the rest of the Krutrim team which helped with model development at various stages. Specifically, we want to thank Anagha Bhangare, Raja Kolla and Aditya Kallappa for the discussions. Our models were trained with generous support from Krutrim cloud using Krutrim credits. We also thank the challenge organizers as well as the reviewers for their valuable feedback and suggestions. 
\bibliography{anthology,custom,copy}
\bibliographystyle{acl_natbib}

\appendix



\end{document}